\pdfoutput=1

\documentclass[11pt]{article}

\usepackage{EMNLP2022}

\usepackage{times}
\usepackage{latexsym}

\usepackage{todonotes}
\usepackage{amsmath} 
\usepackage[shortlabels]{enumitem}
\usepackage{comment}
\usepackage{multirow}
\usepackage{booktabs}
\usepackage{float}
\usepackage{xcolor}

\definecolor{verde}{rgb}{0.0, 0.5, 0.0}

\usepackage[T1]{fontenc}

\usepackage[utf8]{inputenc}

\usepackage{microtype}

\usepackage{inconsolata}

\title{The Undesirable Dependence on Frequency of Gender Bias Metrics Based on Word Embeddings}

\author{
    Francisco Valentini \\ 
        ICC (UBA-CONICET) \\ 
        Maestría en Data Mining (UBA) \\
        Buenos Aires, Argentina \\
        \texttt{ft.valentini@gmail.com} \\
    \And
    Germán Rosati \\
        Escuela IDAES (UNSAM) \\
        Buenos Aires, Argentina \\
        \texttt{grosati@unsam.edu.ar} \\ 
    \AND
    Diego Fernandez Slezak \\
        ICC (UBA-CONICET) \\
        Buenos Aires, Argentina \\
        \texttt{dfslezak@dc.uba.ar} \\
    \And
    Edgar Altszyler \\
        ICC (UBA-CONICET) \\ 
        Maestría en Data Mining (UBA) \\
        Buenos Aires, Argentina \\
        \texttt{ealtszyler@dc.uba.ar} \\
}

\begin{document}
\maketitle
\begin{abstract}
Numerous works use word embedding-based metrics to quantify societal biases and stereotypes in texts. Recent studies have found that word embeddings can capture semantic similarity but may be affected by word frequency. In this work we study the effect of frequency when measuring female vs. male gender bias with word embedding-based bias quantification methods. We find that Skip-gram with negative sampling and GloVe tend to detect male bias in high frequency words, while GloVe tends to return female bias in low frequency words. We show these behaviors still exist when words are randomly shuffled. This proves that the frequency-based effect observed in unshuffled corpora stems from properties of the metric rather than from word associations. The effect is spurious and problematic since bias metrics should depend exclusively on word co-occurrences and not individual word frequencies. Finally, we compare these results with the ones obtained with an alternative metric based on Pointwise Mutual Information. We find that this metric does not show a clear dependence on frequency, even though it is slightly skewed towards male bias across all frequencies.
\end{abstract}

\section{Introduction} \label{sec:introduction}

Word embeddings are one of the most commonly used techniques to measure semantic closeness between words in a corpus. In recent years, they have been widely used in Computational Social Science applications to measure societal biases and stereotypes \citep{caliskan2017semantics,garg2018word,kozlowski2019geometry,lewis2020gender,charlesworth2021gender}.  


For practical purposes, we consider bias to be the degree to which the language used to describe groups or things is different. Bias is typically measured by computing the difference between the mean similarity of words of two context groups $A$ and $B$ with respect to a target word $x$:
\begin{equation} \label{eq:BiasWE} 
    \text{Bias}_{\text{WE}} = 
        \underset{a \in A}{\mathrm{mean}} \; \text{cos}(v_x,v_a) - 
        \underset{b \in B}{\mathrm{mean}} \; \text{cos}(v_x,v_b),
\end{equation}
where $v_i$ is the word embedding of word $i$ and $\text{cos}(v_i,v_j)$ is the cosine similarity between vectors.

Gender bias has long been one of the most studied biases with this method. In this context, $A$ and $B$ are usually defined as gendered nouns and pronouns \citep{caliskan2017semantics,lewis2020gender}. A representative example is \citet{garg2018word}, where the female vs. male bias of professions in historical corpora is found to correlate with the percentage of women employed in each profession over time. 

Not as widely used as word embeddings, Pointwise Mutual Information (PMI) is a metric of word similarity that can also be used to study biases \citep{galvez2018half,aka2021measuring,valentini2021bias}. \citet{valentini2021bias} define the PMI-based bias metric as
\begin{equation*} 
 \text{Bias}_{\operatorname{PMI}} =  
    \operatorname{PMI}(x,A) - \operatorname{PMI}(x,B),
\end{equation*}
where
\begin{equation*}
    \operatorname{PMI}(x,Y) 
        = \log \, \frac{P(x,Y)}{P(x)P(Y)}.
\end{equation*}
$P(x,Y)$ is the probability of co-occurrence between the word $x$ with any one in $Y$ in a window of a predefined number of words, and $P(x)$ and $P(Y)$ are the probability of occurrence of the word $x$ and any word in $Y$, respectively. \citet{valentini2021bias} show that $\text{Bias}_{\operatorname{PMI}}$ can be expressed as 
\begin{equation} \label{eq:BiasPMI}
 \text{Bias}_{\operatorname{PMI}} 
    = \log \, \frac{P(x|A)}{P(x|B)}.
\end{equation}

That is, $\text{Bias}_{\operatorname{PMI}}$ can be interpreted as how much more likely it is to find words in $x$ in the context of words in $A$ than in the context of words in $B$, in a log scale. Thus, it captures exclusively first-order associations and can be computed via maximum likelihood using co-occurrence counts from the text \citep{valentini2021bias}.

Some recent works have studied the relationship between word frequencies and word embeddings. In particular, embeddings seem to encode word frequency even after normalization \citep{schnabel2015evaluation}, vector norm depends on word frequency \citep{wilson2015controlled}, top principal component directions encode frequency in different ways \citep{mu2018allbutthetop} and vectors of high-frequency and low-frequency words lie in different regions of the embedding space \citep{gong2018frage}. 

These studies are nevertheless inconclusive in the sense that they do not determine clearly to what extent the association observed is caused by actual attributes of corpora or by undesirable properties of embedding training. Hence, an answer to the origin of the relation between embeddings and frequency is yet to be found. What is more, the repercussions of this effect in applications relevant to Computational Social Sciences such as bias quantification have not yet been explored. 

We make three main contributions. First, we show that frequency has an association with gender bias when measured with word embedding-based metrics: both Skip-gram with negative sampling (SGNS) and GloVe-based bias metrics tend to detect male bias in high frequency words, while GloVe also yields female bias on average in low-frequency words. Second, we show that the dependence of the embedding-based gender bias on frequency holds when tokens in the corpus are randomly shuffled. This proves that the dependence on frequency is an artifact of the metric itself i.e. that embedding-based bias metrics can encode frequency spuriously. Third, we find that the PMI-based gender bias metric does not present this frequency-based effect but is slightly skewed towards male bias across all frequency ranges.\footnote{Code for the paper is available at \url{https://github.com/ftvalentini/EmbeddingsBiasFrequency}}

Our analyses are restricted to the English language and are based on a binary understanding of gender (see \nameref{sec:limitations}).

\section{The effect of frequency on gender bias} \label{sec:experiments}

Our objective is to study the association between gender bias and frequency in the widely used embedding-based metrics and in the alternative PMI-based metric. Therefore, in a first experiment, we analyze this in two pretrained word embeddings, GloVe \citep{pennington2014glove} and Word2Vec with SGNS \citep{mikolov2013distributed}. 

We do this by studying the distribution of bias in different bins of frequency of words in the vocabulary. Bias is computed with equation \ref{eq:BiasWE} with the female and male context words lists used in \citet{caliskan2017semantics}, and we refer to this as female bias or gender bias. For each frequency bin, we also compute the ratio between the mean and the sample standard deviation (SD). These are Cohen’s $d$ effect sizes of the mean of each group under the null hypothesis of absence of bias on average \citep{cohen1988statistical}. Here we use it as a normalized magnitude of the deviation of the distribution from zero. We use this methodology to assess the association between gender bias and frequency hereinafter. See Appendix \ref{sec:methods} for further details.

\begin{figure}[!ht]
    \centering
    \includegraphics[width=\linewidth]{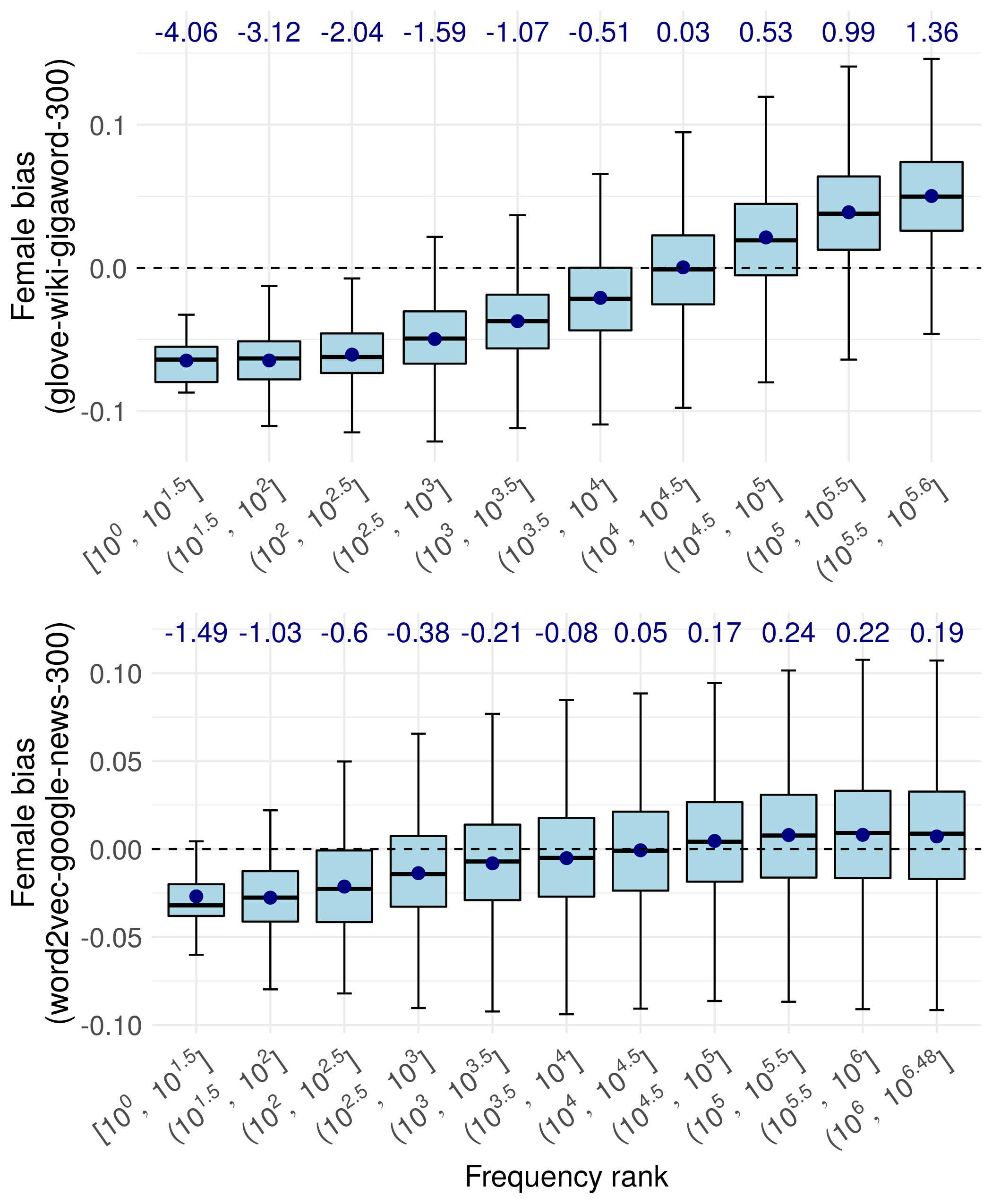}
    \caption{Female bias distribution vs. words' frequency rank in pretrained GloVe (top panel) and Word2Vec with SGNS (bottom panel). Words are grouped into bins according to their rank in a log-scale, so that the most frequent words are in the leftmost bin and the less frequent, in the rightmost. We use frequency ranks as raw frequencies are not available for pretrained embeddings. Blue dots represent the means and blue values are the effect sizes (mean to SD ratio). The plots are not comparable in either axis because the corpus, the vocabulary and the training methodology of each set of embeddings are different.}
    \label{fig:bias_pretrained}
\end{figure}

There is a clear association between gender $\text{Bias}_{\operatorname{WE}}$ with the pretrained embeddings and target word frequency (Figure \ref{fig:bias_pretrained}). GloVe embeddings present a monotonic relationship between frequency and gender bias, such that the top $10^{3.5}$ words tend to have male bias with large effect sizes, whereas less frequent words have mean female bias with medium to large effect sizes. In the SGNS embeddings the effect is small and positive in less frequent words, but in the top 100 words there is a large shift towards male bias. 

Even if there is literature which has studied the relationship between frequency and word vectors (see section \ref{sec:introduction}), this result is still startling: \textit{a priori}, we wouldn't expect the gender bias of words to correlate so strongly with frequency, because word similarity should be more closely related to semantics and co-occurrences in the training corpus than with the individual frequencies of words. 

To validate this behavior, we train GloVe and SGNS embeddings from scratch with the English Wikipedia and study the association between gender bias and word frequency. We compare this with the results obtained with $\text{Bias}_{\operatorname{PMI}}$ (equation \ref{eq:BiasPMI}).

\subsection{Comparing $\text{Bias}_{\operatorname{WE}}$ with $\text{Bias}_{\operatorname{PMI}}$} \label{sec:bias_unshuffled}

\begin{figure}[!ht]
    \centering
    \includegraphics[width=\linewidth]{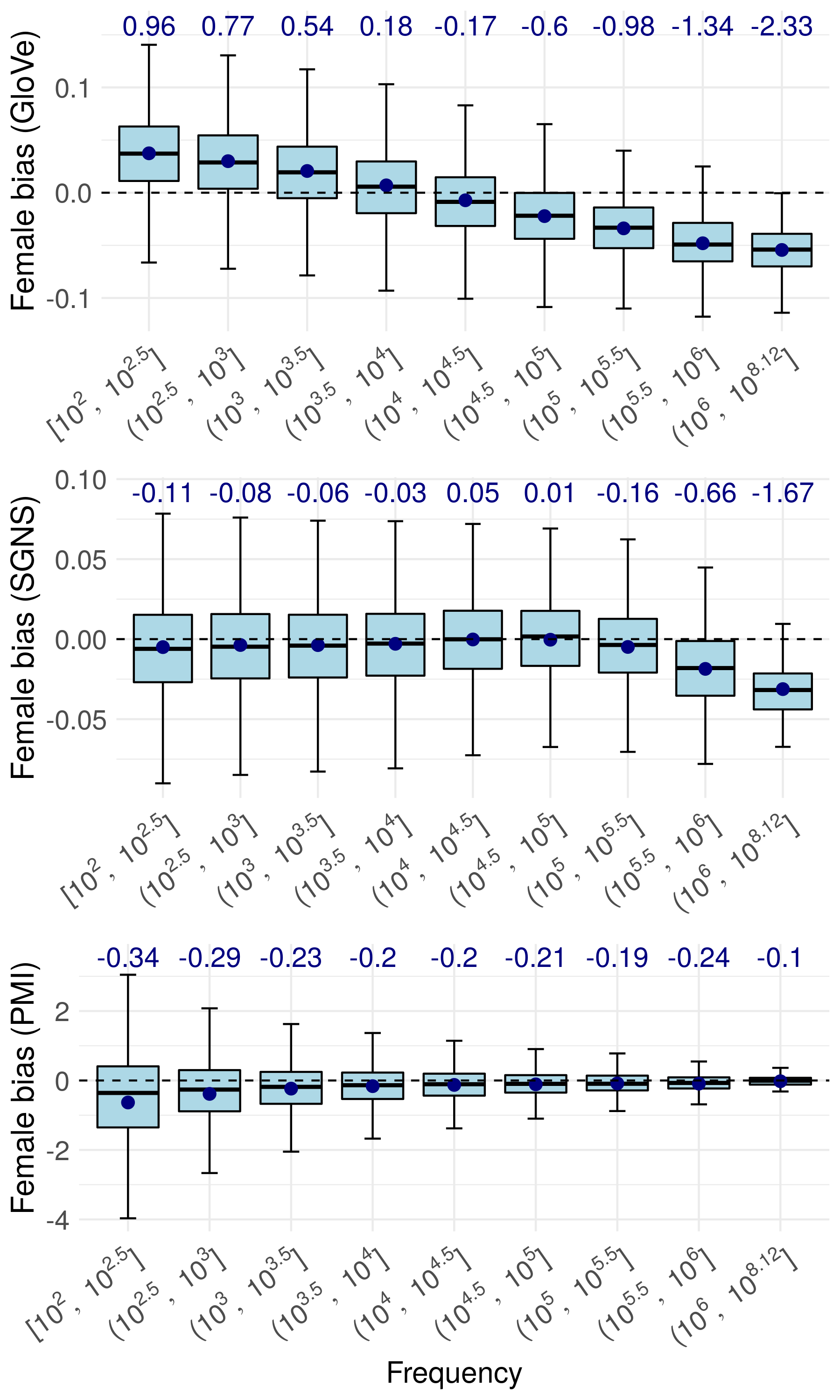}
    \caption{Female bias vs. frequency in Wikipedia. Bias is measured with $\text{Bias}_{\operatorname{WE}}$ using GloVe (top panel), $\text{Bias}_{\operatorname{WE}}$ using SGNS (middle panel), and $\text{Bias}_{\operatorname{PMI}}$ (bottom panel). Words in the vocabulary are grouped in bins according to their frequencies in log-scale. Blue dots represent the means and blue values are the effect sizes (mean to SD ratio).}
    \label{fig:bias_unshuffled}
\end{figure}

\begin{description}[wide,itemindent=\labelsep]

\item[Methods and data] We measure the gender bias of words in the vocabulary of the 2021 English Wikipedia with $\text{Bias}_{\operatorname{PMI}}$ and $\text{Bias}_{\operatorname{WE}}$ and assess the association with word frequency. We train SGNS and GloVe vectors to compute $\text{Bias}_{\operatorname{WE}}$, whereas the frequency counts from the corpus are used to compute $\text{Bias}_{\operatorname{PMI}}$. Refer to appendices \ref{sec:corpus} and \ref{sec:methods} for details on the corpus and the methods, respectively.

\item[Results] The relation between $\text{Bias}_{\operatorname{WE}}$ and frequency in pretrained embeddings (Figure \ref{fig:bias_pretrained}) holds qualitatively when training embeddings from scratch (top and middle panels in Figure \ref{fig:bias_unshuffled}). GloVe embeddings yield a negative relationship between female bias and frequency, while in SGNS we find an average male bias with medium to large effect sizes in high frequency words.

When using $\text{Bias}_{\operatorname{PMI}}$ no frequency bin is strongly biased on average (bottom panel in Figure \ref{fig:bias_unshuffled}). There is however a slight skew towards male bias, such that all frequency ranges present small negative effect sizes. Furthermore, the variability of bias tends to increase as the frequency of target words decreases: this behavior is attributable to the fact that PMI is usually high and noisy in low frequency words \citep{jurafsky2000speech}.

This analysis is not enough to determine that the effect of frequency on embedding-based bias metrics is a spurious artifact generated by the embeddings. It still might be the case that higher frequency words are actually more male-biased than lower frequency words due to second-order or higher associations, thus yielding plots like those on the top and middle panels of Figure \ref{fig:bias_unshuffled}. We conduct the following study to assess this.

\end{description}

\subsection{The undesirable dependency on frequency} \label{sec:bias_shuffled}

\begin{figure}[!ht]
    \centering
    \includegraphics[width=\linewidth]{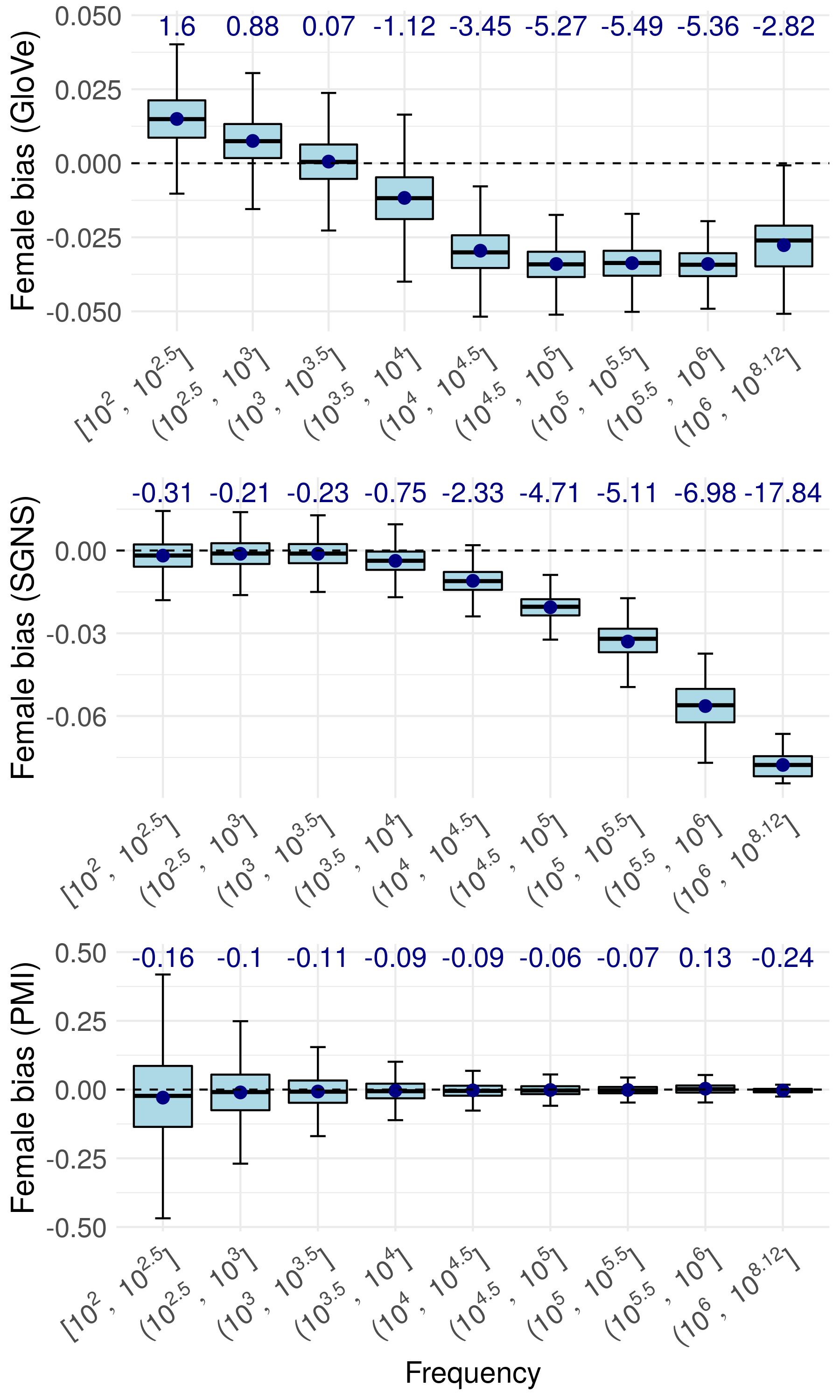}
    \caption{Female bias vs. frequency in shuffled Wikipedia. The bias of each word is computed as the average of five estimates, one for each of five shuffles performed. Words are grouped in bins according to their frequencies. Blue dots represent the means and blue values are the effect sizes (mean to SD ratio).}
    \label{fig:bias_shuffled}
\end{figure}

\begin{description}[wide,itemindent=\labelsep]

\item[Methods and data] We create five randomly shuffled, independent versions of the Wikipedia corpus where tokens are randomly located across the text. In these corpora words keep their frequency but lose their context because co-occurrences are completely random. We estimate bias with $\text{Bias}_{\operatorname{WE}}$ and $\text{Bias}_{\operatorname{PMI}}$ in each of the corpora and consider the average of the five values as the gender bias of each word. We analyze the relationship between the gender bias metrics and frequency in this setting. By shuffling the words in the corpus, contexts become meaningless, thus any association found between bias and frequency in this setting is explained only by the frequencies of the words. We highlight that it is problematic and undesirable for a metric to detect biases in a setting where they do not exist. 

\item[Results] In this controlled experimental setup, $\text{Bias}_{\operatorname{WE}}$ presents a strong association with target word frequency (Figure \ref{fig:bias_shuffled}): average male bias grows as frequency increases for both SGNS and GloVe, with large effect sizes from around frequencies $10^4$ onwards. Low frequency words present female bias on average when measured with GloVe, while they tend to have a slight male bias with small effect sizes when using SGNS. 

Conversely, measuring bias with PMI in the shuffled corpora does not produce a clear dependence on frequency. The average bias is roughly constant for all frequencies, with small negative effect sizes; that is, there is a slight skew towards male bias across all frequencies.

\end{description}

\section{Discussion and Conclusion} \label{sec:conclusion}

In this work we revealed the existence of a spurious frequency-based distortion in gender bias metrics based on the cosine similarity between word embeddings. Both SGNS and GloVe-based gender bias metrics tend to detect male bias in high frequency words, while GloVe also yields female bias on average in low frequency words. 

To determine whether this effect is indeed an undesirable artifact of the embedding-based metric we assessed the relation between gender bias and frequency in shuffled corpora, where words lose their context but keep their frequency. Results reveal that the dependence on frequency is caused by the metric and does not originate from actual properties of the texts. This shows that popular gender bias measurements can detect bias even when there is none. Additionally, we found that an alternative PMI-based bias metric does not show a clear dependence on frequency, even though it shows a slight tendency towards male bias.

According to these results, we consider the PMI-based bias metric has an advantage over the embedding-based metrics, which adds to the advantages of interpretability and hypothesis testing \citep{valentini2021bias}. However, as PMI captures exclusively first-order associations and is unable to capture synonyms, 
it may be required to include several terms associated to the context words in order to measure some biases.

Male nouns and pronouns are usually more frequent than female ones in large corpora \citep{twenge2012male, galvez2018half}. For example, in the Wikipedia corpus, \textit{he} appears 11.8 million times, while the frequency of \textit{she} is 3.5 million (refer to Appendix \ref{sec:methods} for the frequencies of the other gendered context words). 

The disparity in frequencies of male and female contexts is a type of bias in itself and can be measured by counting word occurrences. In contrast, the bias we study refers to the stereotyped contexts in which male and female entities are portrayed, and should be independent of individual word frequencies.


When words are shuffled, the biases associated with the contexts of female and male context words are eliminated, but the disparities in frequencies are maintained. We propose that bias metrics capture this disparity in frequencies of female and male context words. In the case of the embedding-based metric, this hypothesis is supported by the existing evidence that embeddings encode word frequency in addition to semantics.

We believe the random-shuffling experiment is general enough to show that the frequency effect would still exist with other word lists, types of biases and domains, as long as the frequencies of the context words differ. This result is important because the context words’ frequencies are disregarded when measuring biases with embeddings.

Our findings have important implications for bias measurement applications, as they cast doubt on the reliability of widely used bias metrics when the frequencies of the words involved are very different. We believe that more effort should be put into designing new bias detection methods that do not suffer from this weakness.

\section*{Limitations} \label{sec:limitations}


We use sets of context words typically used in the gender bias literature. These words imply a binary understanding of gender, excluding other gender representations from the bias measurement. Moreover, we focus exclusively on the English Wikipedia corpus and do not apply methods on corpora of other domains, which might yield different distributions of gender bias. 

We report results using default hyperparameters. This intends to mimic the typical experimental setup found in the Computational Social Science literature. Hyperparameters are left at their default values because there is no ground truth for biases, i.e. there are no annotations indicating the level of bias of words.

The studies conducted in this work can be adapted to other languages, other biases and other corpora. We hope further research can assess the frequency-based distortion in these settings as well as the influence of hyperparameter choices.

\bibliographystyle{acl_natbib}
\bibliography{anthology,custom}

\appendix

\section{Corpus} \label{sec:corpus}

We use the April 2021 Wikipedia dump\footnote{\url{https://archive.org/download/enwiki-20210401}} and remove articles with less than 50 tokens. We remove non-alpha-numeric symbols and apply sentence splitting. The corpus contains around 1.7 billion tokens and 78.1 million documents (sentences) after pre-processing.

\section{Methods} \label{sec:methods}

We measure female vs. male gender using gendered nouns and pronouns \citep{caliskan2017semantics,lewis2020gender}, namely, A=\textit{\{female, woman, girl, sister, she, her, hers, daughter\}} and B=\textit{\{male, man, boy, brother, he, him, his, son\}}. 

Tables \ref{tab:female_frequencies} and \ref{tab:male_frequencies} display the frequency of each of these words in the pre-processed Wikipedia corpus.

\begin{table}[H]
\centering
\begin{tabular}{lr}
\toprule
Word & Frequency\\
\midrule
her & 3,720,408\\
she & 3,517,570\\
daughter & 294,043\\
female & 282,159\\
woman & 236,954\\
sister & 179,511\\
girl & 141,616\\
hers & 5,706\\
\bottomrule
\end{tabular}
\caption{Frequencies of female context words in the Wikipedia corpus}
\label{tab:female_frequencies}
\end{table}

\begin{table}[H]
\centering
\begin{tabular}{lr}
\toprule
Word & Frequency\\
\midrule
he & 11,815,189\\
his & 9,603,118\\
him & 1,811,552\\
son & 541,828\\
man & 443,881\\
brother & 287,544\\
male & 181,471\\
boy & 124,326\\
\bottomrule
\end{tabular}
\caption{Frequencies of male context words in the Wikipedia corpus}
\label{tab:male_frequencies}
\end{table}

We exclude words with fewer than 100 occurrences, which yields a vocabulary of 222,144 words. Table \ref{tab:frequency_distribution} displays the distribution of these words according to their frequencies, excluding the female and male context words. 

\begin{table}[H]
\centering
\begin{tabular}{lr}
\toprule
Frequency & \# words\\
\midrule
$[10^{2},10^{2.5}]$ & 116,340\\
$(10^{2.5},10^{3}]$ & 54,187\\
$(10^{3},10^{3.5}]$ & 26,617\\
$(10^{3.5},10^{4}]$ & 13,144\\
$(10^{4},10^{4.5}]$ & 6,579\\
$(10^{4.5},10^{5}]$ & 3,255\\
$(10^{5},10^{5.5}]$ & 1,448\\
$(10^{5.5},10^{6}]$ & 441\\
$(10^{6},10^{8.12}]$ & 117\\
\bottomrule
\end{tabular}
\caption{Number of words in each frequency range in the Wikipedia corpus}
\label{tab:frequency_distribution}
\end{table}

In section \ref{sec:experiments} we use pretrained GloVe embeddings trained on Wikipedia 2014 and Gigaword 5 \citep{pennington2014glove}, and Word2Vec SGNS embeddings trained on Google News \citep{mikolov2013distributed}, both with 300 dimensions. 

All methods employed in sections \ref{sec:bias_unshuffled} and \ref{sec:bias_shuffled} (GloVe, SGNS and PMI) use a window size of 10 and remove out-of-vocabulary tokens before the corpus is processed into word-context pairs \citep{levy2015improving}.

For SGNS we use the Word2Vec implementation available in the Gensim library \citep{rehurek2010gensim} with default hyperparameters. GloVe is trained with \citet{pennington2014glove}’s implementation with 100 iterations. 

For PMI, we count co-occurrences with the GloVe module \citep{pennington2014glove} and set the smoothing parameter $\epsilon$ to 0.01, so that it can be computed whenever there are no co-occurrences between the target word and any of the context words. 

All computations were performed on a desktop machine with 4 cores Intel Core i5-4460 CPU @ 3.20GHz and 32 GB RAM. Training took around 30 minutes per iteration with GloVe and 2 hours per epoch with SGNS.











\end{document}